\documentclass[lettersize,journal]{IEEEtran}
\usepackage{amsmath,amsfonts}
\usepackage{algorithmic}
\usepackage{algorithm}
\usepackage{array}
\usepackage[T1]{fontenc}
\usepackage{textcomp}
\usepackage{stfloats}
\usepackage{url}
\usepackage{verbatim}
\usepackage{graphicx}
\usepackage{cite}
\usepackage{pifont}
\usepackage{bbm}
\usepackage{amsmath}
\usepackage{amssymb}
\usepackage{booktabs}
\usepackage{multirow}
\usepackage{subfigure} 
\usepackage{makecell}
\usepackage{xcolor}

\newcommand{\eg}{\emph{e.g.,}~}

\newcommand{\ie}{\emph{i.e.,}~}

\usepackage[capitalize]{cleveref}
\crefname{section}{Sec.}{Secs.}
\crefname{table}{Tab.}{Tabs.}

\begin{document}

\title{Dual-view Curricular Optimal Transport for Cross-lingual Cross-modal Retrieval}

\author{Yabing Wang, Shuhui Wang, Hao Luo, Jianfeng Dong, Fan Wang, Meng Han, Xun Wang, and Meng Wang
\thanks{Y. Wang, J. Dong and X. Wang are with the College of Computer Science and Technology, Zhejiang Gongshang University, Hangzhou 310035, China.}
\thanks{S. Wang is with the Institute of Computing Technology, Chinese Academy of Sciences, Beijing 100190, China.}
\thanks{H. Luo and F. Wang are with Alibaba Group, Hangzhou 310052, China.}
\thanks{M. Han is with the Binjiang Institute of Zhejiang University,
Hangzhou 310027, China.}
\thanks{M. Wang is with the School of Computer Science and Information
Engineering, Hefei University of Technology, Hefei 230009, China.}

\thanks{Manuscript received June 19, 2021; revised X, 2023.}}

\markboth{Journal of \LaTeX\ Class Files,~Vol.~X, No.~X, June~2023}%
{Shell \MakeLowercase{\textit{et al.}}: A Sample Article Using IEEEtran.cls for IEEE Journals}

\IEEEpubid{0000--0000/00\$00.00~\copyright~2023 IEEE}

\maketitle

\begin{abstract}
Current research on cross-modal retrieval is mostly English-oriented, as the availability of a large number of English-oriented human-labeled vision-language corpora. 
In order to break the limit of non-English labeled data, cross-lingual cross-modal retrieval (CCR) has attracted increasing attention.
Most CCR methods construct pseudo-parallel vision-language corpora via Machine Translation (MT) to achieve cross-lingual transfer. 
However, the translated sentences from MT are generally imperfect in describing the corresponding visual contents.
Improperly assuming the pseudo-parallel data are correctly correlated will make the networks overfit to the noisy correspondence.
Therefore, we propose Dual-view Curricular Optimal Transport (DCOT) to learn with noisy correspondence in CCR. 
In particular, we quantify the confidence of the sample pair correlation with optimal transport theory from both the cross-lingual and cross-modal views, and design dual-view curriculum learning to dynamically model the transportation costs according to the learning stage of the two views.
Extensive experiments are conducted on two multilingual image-text datasets and one video-text dataset, and the results demonstrate the effectiveness and robustness of the proposed method.
Besides, our proposed method also shows a good expansibility to cross-lingual image-text baselines and a decent generalization on out-of-domain data.

\end{abstract}


\section{Introduction}
\IEEEPARstart{C}{ross-lingual} Cross-modal Retrieval (CCR) retrieves the visual contents (\ie videos or images) which are semantically relevant based on target-language (\eg non-English) queries $T$, but can only be trained on the manually annotated pairs of visual contents $V$ and source language (\eg English) captions $S$.
It aims to alleviate the problem of the existence of large-scale multilingual vision-language corpora and the limited development of non-English languages in the field of cross-modal retrieval \cite{shvetsova2022everything, han2022temporal, ge2022bridging, qi2021semantics, dong2022reading, li2021memorize, zhang2020deep, zhang2022latent, qin2022joint, dong2022dual}.

\begin{figure}[tb!]
\centering\includegraphics[width=0.9\columnwidth]{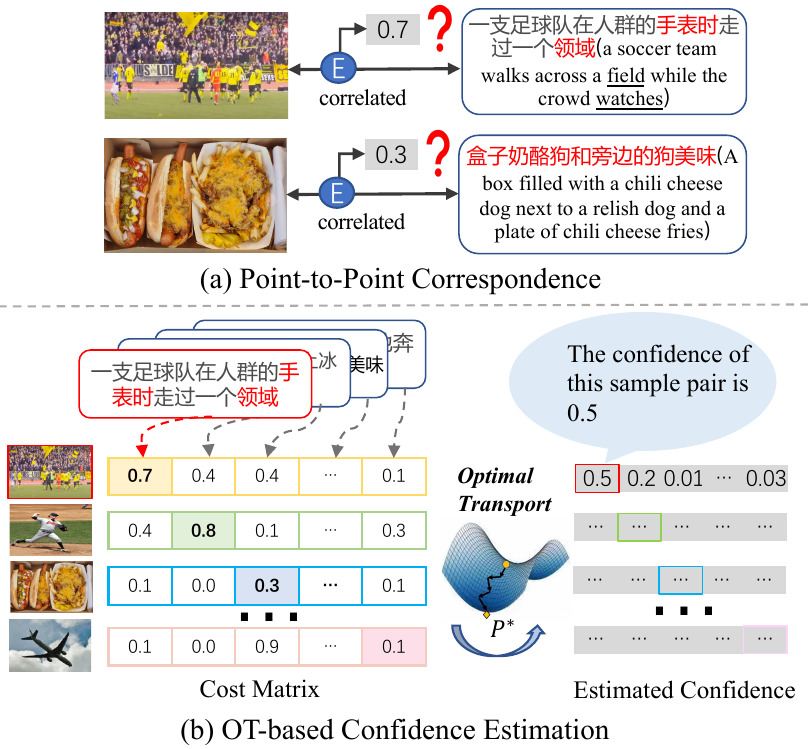}
\caption{An example of the P2P-based method and our proposed OT-based confidence estimation method. The red words represent the incorrectly translated ones.
 }\label{fig:title_pic}
 \vspace{-4mm}
\end{figure}

The key of CCR is how to achieve effective cross-lingual transfer to facilitate alignment between visual and target-language features.
Recently, a series of breakthroughs have been proposed \cite{aggarwal2020towards, lei2021mtvr, portaz2019image, zhou2021uc2, huang2021multilingual, fei2021cross, ni2021m3p, wang2022cross, zeng2022cross}.
Instead of relying on the parallel corpus for direct visual-target language alignment, some works~\cite{ni2021m3p, fei2021cross} utilize source language as the focal point to build a bridge between visual content and target language.
However, they fail to break the semantic gap between the visual and target language and the parallel corpus is still costly to collect. 
With the popularity of Machine Translation (MT), a natural solution~\cite{zhou2021uc2, zeng2022cross} is to generate pseudo visual and target language pairs by MT and directly establish their correspondences.
In specific, \cite{zhou2021uc2, zeng2022cross} pre-train the model with a large number of pairs of visual data and translated target-language captions (V+T). However, they still rely on large-scale vision-language datasets (\emph{e.g.,} CC3M~\cite{sharma2018conceptual} and its translation) and ignore the noise from translation. As shown in \cref{fig:title_pic}~(a), even with the most powerful off-the-shelf MT tools, the translated target-language captions still contain various noises, such as spelling errors, grammar errors, and even distorted overall meaning.

\IEEEpubidadjcol
Due to the noise introduced during the translation process, the imperfect target-language captions cannot accurately describe the corresponding visual contents (\ie noise correspondence problem).
In this case, if we persist in promoting the alignment between the visual and target-language features in a common space, the model will overfit to the wrong supervision and result in degraded performance.
A recent method called Noise-Robust Cross-lingual Cross-modal Retrieval (NRCCR)~\cite{wang2022cross}
employs multi-view distillation to generate soft pseudo-targets as direct supervision for target-language learning, and achieves comparable results to methods of using extra pre-training data.
Considering the ubiquitousness of noisy correspondence in various cross-domain matching tasks, there have been consistent endeavors to alleviate its adversarial influence.
A typical solution is to use Cosine or Euclidean distance as the point-to-point~(P2P) correspondence of sample pairs \cite{li2021align, han2022temporal}.
However, similar with NRCCR, these methods (\cref{fig:title_pic}~(a)) 
ignore the instance relation in context points in two sets of data, while the context information between sample pairs is crucial for reliable correspondence.
Some works~\cite{huang2021learning,yang2022learning} utilize the Gaussian Mixture Model (GMM) to divide the data into clean and noisy partitions. 
Although GMM considers the distribution relation between samples, the assumption of mixture of Gaussian may not capture complex and diverse patterns in real-world data (\eg long-tail distribution).
Moreover, these methods only address the matching problem between two sets of instances ($V \leftrightarrow T$ or $S \leftrightarrow T$), which is not well-suited for CCR with multiple domains ($V\leftrightarrow T \leftrightarrow S$).

To tackle the aforementioned limitations, this paper proposes a CCR-specific noise-robust method called Dual-view Curricular Optimal Transport (DCOT).
To obtain reliable correspondence, we formulate the noisy correspondence learning as an optimal transport (OT) problem.
Instead of finding correspondences between individual points of two sets, our method aims to find an optimal matching between the two sets (\cref{fig:title_pic}~(b)).
We interpret the optimal matching as a confidence measure for the correct matching between sample pairs, which allows us to evaluate the reliability of the correlation score between sample pairs.
Considering the presence of multiple domains in CCR, we incorporate both cross-lingual and cross-modal views and use OT from both views to quantify the confidence of each correlated sample pair.
Through theoretical and empirical analysis, we found that the model fitting is undertaken quickly on cross-lingual view in the early stage, and gradually transferred to cross-modal view in the later stage.
Accordingly, we design a dual-view curriculum learning process, 
which constructs the transportation costs and determines the weights of both views dynamically with a curriculum schedule based on the learning status of the two views at each time-step during training.
Our method is more flexible to different types of noise in CCR, as we do not make any assumptions about the underlying data distribution.
Our contributions can be summarized as follows:
\begin{itemize}
    \item To take into account the instance relation in context, we formulate the noisy correspondence learning in CCR as an optimal transport problem.
    \item The proposed DCOT method dynamically models the transportation costs according to the learning state of two views, \ie cross-lingual and cross-modal views, to avoid overfitting to the noisy sample pairs.
    \item Extensive experiments on three image-text and video-text cross-modal retrieval benchmarks across different languages  demonstrate the effectiveness and robustness of our method.
\end{itemize}

\begin{figure*}[tb!]
\centering\includegraphics[width=2.0\columnwidth]{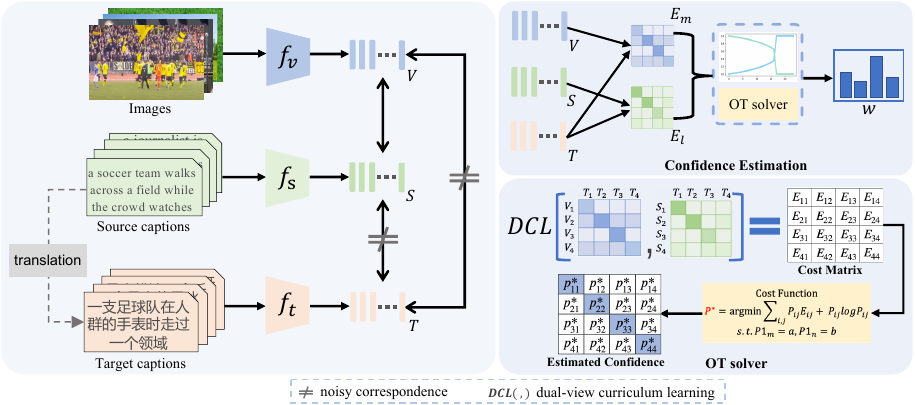}
\caption{
\textbf{Illustration of our proposed DCOT method for CCR.}
The images, source-language captions and target-language captions are first encoded to representations. Then the confidence of the correctly correlated for $(V, T)$ pairs is estimated by OT-based confidence estimation from both the cross-lingual view and cross-modal views.
For confidence estimation, we design a dual-view curriculum learning, by which the transportation costs modeling is tied to the learning state of the two views. 
Finally, DCOT adjusts the contribution of each pair dynamically according to the estimated confidence during training.
}\label{fig:framework}
\vspace{-3mm}
\end{figure*}

\section{Related Works}

\subsection{Cross-lingual Cross-modal Retrieval}

Cross-lingual transfer learning has become a crucial mechanism to battle the unavailability of annotated low-resource languages. 
Recently, some works \cite{aggarwal2020towards, lei2021mtvr, portaz2019image, zhou2021uc2, huang2021multilingual, fei2021cross, ni2021m3p, wang2022cross} try to apply cross-lingual transfer learning to cross-modal retrieval tasks to alleviate the problem of data scarcity and achieved remarkable progress. 
Under the CCR setting, the model trained on manually annotated pairs of vision and source language is adapted for evaluations in different target languages.
Prior works \cite{portaz2019image, aggarwal2020towards} aligning different languages into a common space with non-contextualized multilingual word embeddings (MUSE \cite{conneau2017word} and BIVEC \cite{luong2015bilingual}) and pre-trained sentence encoders(mUSE \cite{yang2019multilingual} and LASER \cite{conneau2017word}), respectively.
The major study can be divided into three groups based on how the alignment of visual-target language is achieved: 
1) rely on parallel corpus, 2) collect multilingual subtitles from the web, and 3) resort to MT.

To be specific, methods that rely on parallel corpus \cite{ni2021m3p, fei2021cross} regard English as the focal point to build a bridge between visual and target languages. For example, the code-switched training\cite{ni2021m3p} enforces the explicit alignment between images and non-English languages.
However, these methods indirectly align visual and non-English languages centered on English, and the process of collecting parallel corpus is also costly and time-consuming.
Huang {\it et al.}~\cite{huang2021multilingual} crawl and collect the multilingual subtitles from YouTube, and extend the HowTo100M \cite{miech2019howto100m} to the multilingual version. 
Multilingual Vision-Language data corpus with MT
\cite{zhou2021uc2, wang2022cross, zeng2022cross} have recently emerged as an alternative. 
An MT-augmented cross-lingual cross-modal pretraining framework~\cite{zhou2021uc2} is proposed, which pivots primarily on images and complementarily on English for multilingual multi-modal representation learning. Further, a noise-robust learning framework~\cite{wang2022cross} is proposed to deal with the  noise in MT results. 
However, they do not explicitly evaluate the confidence of the samples but filter the noise in the target language implicitly by introducing a cross-attention module. This may lead to a suboptimal solution for solving the model overfitting in the presence of noisy labels.

\subsection{Learning with Noisy Correspondence} 
The issue of noisy labels has been well studied in the visual classification task~\cite{tan2021co, wei2020combating, kim2019nlnl}.
For multi-modal learning, a new paradigm\cite{huang2021learning} is developed, which considers the alignment errors in paired data instead of the errors in category annotations in the classification task.
Recently, some research focused on noise correspondence, such as \cite{huang2021learning, han2022temporal, hu2021learning, li2021align，hu2023cross} in cross-modal retrieval task, and \cite{wang2022cross} in cross-lingual cross-modal retrieval task.
Among them, some works resort to robust architecture design \cite{han2022temporal, fu2022large, wang2022cross, li2021align}. For example,
\cite{wang2022cross} employs multi-view self-distillation to generate soft pseudo-targets to provide direct supervision for noise-robust target-language representation learning.
Other works attempted to evaluate the confidence of sample pairs and design noise-robust losses \cite{huang2021learning, yang2022learning, hu2021learning}. For example, in
~\cite{huang2021learning, yang2022learning}, the annotation confidence among the samples is evaluated  by introducing the GMM and the soft margin of sample pairs adjusted in the triplet loss.
Different from these methods, our method is more generalized and does not rely on any prior information on the input data distribution. It can flexibly combine the cross-lingual and the cross-modal view to estimate the confidence of the correlated sample pairs.

\section{Method}
\subsection{Preliminaries}
Let $\mathcal{D} = \{(V_i, S_i) \}^N_{i=1}$ be a dataset of annotated paired images/videos and source-language captions with data size $N$.
As the access to human-labeled vision and target language sample pairs during training is unavailable, some external tools can be utilized, \eg, MT or parallel corpus.
Following~\cite{wang2022cross}, we extend the training data $\mathcal{D}$ to $\hat{\mathcal{D}} = \{(V_i, S_i, T_i) \}^N_{i=1}$ with MT, where $T_i$ is the translated target-language caption corresponding to $S_i$. 
We define $f_v(.), f_s(.) $ and $f_t(.)$ as the image/video, source-language and target-language encoder, respectively, and denote the embedded fixed-dimensional vectors of an image/video as $V_i$, a source-language caption as $S_i$, and a target-language caption as $T_i$. 
Note that we use the human-input target-language sentences as queries for retrieval during inference.
The framework is illustrated in \cref{fig:framework}.
In what follows,
we will first introduce our proposed OT-based confidence estimation (\cref{sec:confidence-estimation}), then describe the dual-view optimal transportation costs modeling strategy (\cref{sec:curriculum_learning_1}), and finally introduce the noise-aware alignment objective (\cref{sec:noise-aware-objective}).

\subsection{OT-based Confidence Estimation}
\label{sec:confidence-estimation}

{Supposing we have a mini-batch with $M$ image/video-caption pairs, denotes $\overline{\mathcal{D}} = (V,S,T)$. 
Current methods usually utilize the Cosine distance or Euclidean distance to calculate the point-to-point correspondence between individual $(V_i, T_i)$ pair without confidence estimation, which can be expressed as:

\begin{equation}
\label{eq:evaluation_matrix}
\begin{array}{cc}
E = dist(f_v(V), f_t(T)) \in \mathbb{R}^{M \times M}
\end{array}
\end{equation}
where $dist(\cdot)$ is the distance function, \eg, cosine distance, and $E_{ij}$ denotes the correlation score between the $i$-th visual feature and $j$-th translated caption feature.
}

\begin{figure}[tb!]
\centering\includegraphics[width=0.98\columnwidth]{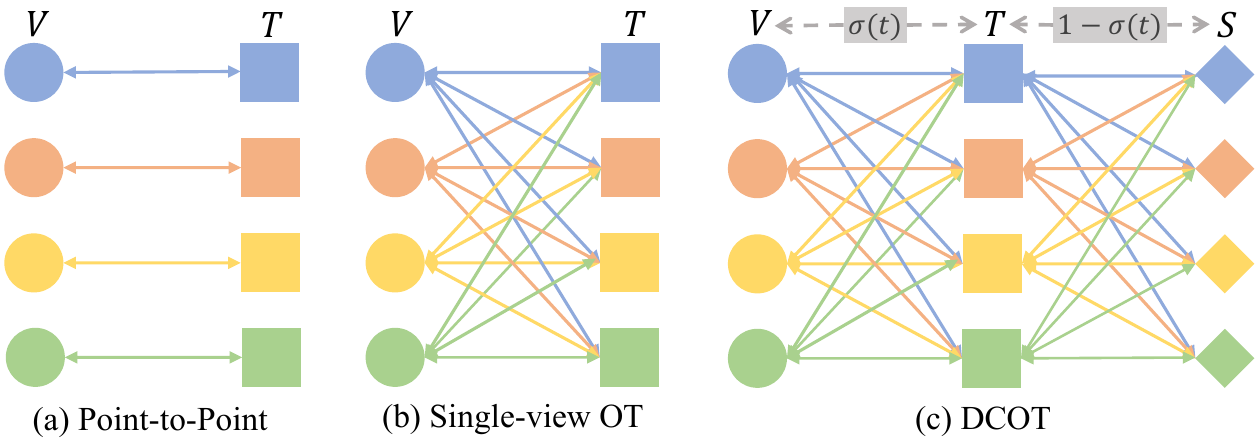}
\caption{
An example of different correspondence calculation methods.
The Point-to-Point method (\eg Cosine distance) does not take into account the contextual relationship between pairs of data. The Single-view OT method estimates the confidence based only on the cross-modal view.
In contrast, our proposed method, DCOT, estimates confidence collaboratively from two views.
}\label{fig:example_OT}
\end{figure}

As illustrated in~\cref{fig:example_OT} (a), the pairwise correlation score of $(V_i, T_i)$ is computed individually without context.
Since the network parameters change dynamically during the training process, the Cosine distance (or Euclidean distance) of the same feature pair at different time steps turns out to be different.
Such a point-to-point correspondence calculation scheme fails to consider the semantic context, and thus it can hardly provide reliable supervision signals, especially under the situation of high noise in the translated sentences. Therefore, to alleviate the noise effect, a better noisy correspondence learning strategy is to consider the mutual relation between features in a data batch to seek for a more reliable contextual correspondence solution.

To address this issue, we propose a solution by formulating noisy correspondence learning as an OT problem and finding the optimal match between data points from two sets.
This enables us to capture the correspondence between the sets from a contextual perspective  (\cref{fig:example_OT}~(b)), and to estimate the confidence of the correlated sample pairs.
Specifically, when the transportation cost between a sample pair is relatively small, their matching degree is higher, indicating that they are highly correlated. Therefore, we can treat the similarity score of the optimal match as a confidence measure for the correct matching between sample pairs. If the similarity score of the optimal match is high, we can consider the matching between sample pairs to be reliable.

\textbf{Optimal transport problem.} 
The Optimal Transport aims to search the most efficient transport plan of transforming one mass distribution to another whilst minimizing the cost.
Specifically, given two discrete point sets, $X = \{x_i\}^n_{i=1}$ and $Y = \{y_j\}^m_{j=1}$, $x_i, y_i \in \mathbb{R}^d$. The amount of mass on these points is given by $a$ and $b$, defined on probability space $X, Y \in \Omega$, respectively.
\begin{equation}
    \begin{array}{cc}
    a = \sum_{i=1}^{n} p_i^x \delta (x_i),b = \sum_{j=1}^{m} p_j^y \delta (y_j)
    \end{array}
\end{equation}
where $\delta (\cdot)$ denotes the Dirac function, $p_i^x$ and $p_j^y$ are the probability mass to the $i$-th and $j$-th sample, belonging to the probability simplex, \ie, $\sum^n_{i=1} p_i^x = \sum^m_{j=1}p_j^y = 1$. 
The unit transportation cost from point $x_i$ to $y_j$ is denoted by $C_{ij}$. 
Under such a setting, we aim to search the optimal transport plan $P^*$ to transport the mass in probability measure $a$ to $b$ with the minimum costs by solving the following problem:
\begin{equation}
\label{ot}
    \begin{array}{cc}
    P^* = \underset{P \in \mathbb{R}^{n \times m}}{\arg\min} \sum_{i,j} P_{ij}C_{ij}  \vspace{1ex} \\
     \text{s.t.} \ P \mathbbm{1}_m = a, \ P^{T} \mathbbm{1}_n = b
    \end{array}
\end{equation}
where $P$ is the transport plan containing all non-negative $n \times m$ elements with row and column sums to $a$ and $b$, respectively.

We define $E$ as the transportation costs denoting transporting one unit of translated caption $T_i$ to image/video $V_i$.
Besides, we add an entropic regularization to control the smoothness of the transport plan following \cite{cuturi2013sinkhorn}. Our goal is to maximize the total correlation to get an optimal estimated confidence on this batch. The corresponding optimization problem can be formulated as follows:

\begin{equation}
\label{equation: our_ot_1}
\begin{array}{cc}
\underset{P \in \mathbb{R}^{M \times M}}{\min} \sum_{i,j} P_{ij}E_{ij} + \frac{1}{\lambda_{reg}} P_{ij}logP_{ij}  \vspace{0.5ex} \\
\text{s.t.} \ P \mathbbm{1}_M = \frac{1}{M} \mathbbm{1}_M , \ P^{T} \mathbbm{1}_M = \frac{1}{M} \mathbbm{1}_M \\
\end{array}
\end{equation}
where $\lambda_{reg} > 0$ is the regularization parameter, a larger $\lambda_{reg}$ leads to ``softer" distribution for $P$, and vice versa.
Note that the constraint condition ensures that the solution satisfies that all instances in the batch are equally important and should be matched with equal probabilities. 
The optimal estimated confidence $P^{*} \in \mathbb{R} ^{M \times M}$ of \cref{equation: our_ot_1} is:

\begin{equation}
\label{eq:get_optimal}
\begin{array}{cc}
P^{*} = diag(\mu) K diag(v) \vspace{1ex}\\
K = \exp( {- \lambda_{reg} E})
\end{array}
\end{equation}
where $u$ and $v$ are some non-negative vectors, solved with Sinkhorn’s fixed point iteration:

\begin{equation}
\label{eq:iter}
\begin{array}{cc}
u^{(t+1)} = \frac{\mathbbm{1}_M}{Kv^{(t)}}, \ v^{(t+1)} =  \frac{\mathbbm{1}_M}{K^T u^{(t+1)}}
\end{array}
\end{equation}
Finally, we take the diagonal element $w_i$ of $P^*$ as the confidence of each $(V_i, T_i)$ pair:
\begin{equation}
    w = diag(P^*)
\end{equation}

\subsection{Dual-view Transportation Cost}
\label{sec:curriculum_learning_1}
As we know, transportation cost is a crucial factor in computing the optimal transport plan.
In this section, we introduce a dual-view curriculum learning approach to dynamically model the transportation costs based on the learning state of two views.

Specifically, in CCR, given a $(V_i,S_i,T_i)$ triplet, $(V_i, S_i)$ represents the ground-truth corresponded pair.
The correspondence between $(V_i, T_i)$ pair can also be inferred from the correspondence between $(S_i, T_i)$ pair.
To obtain more accurate transportation costs,
we calculate them from two views, namely the cross-lingual view ($S\leftrightarrow T$) and the cross-modal view ($V \leftrightarrow T$), respectively:

\begin{figure}[tb!]
\centering
\subfigure[Basic Model]{
\label{fig:dual-view-loss}
\includegraphics[width=0.44\columnwidth]{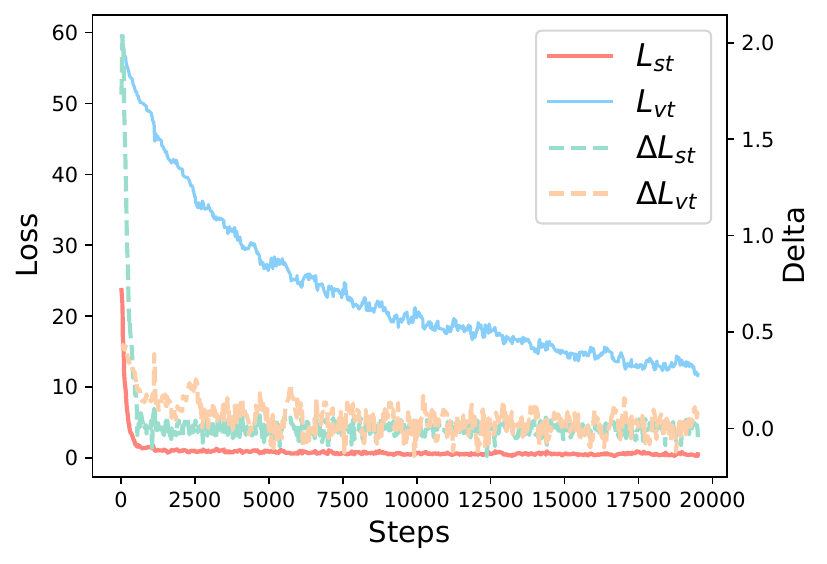}
}
\quad
\subfigure[Ours]{
\label{fig:dual-view-performance}
\includegraphics[width=0.42\columnwidth]{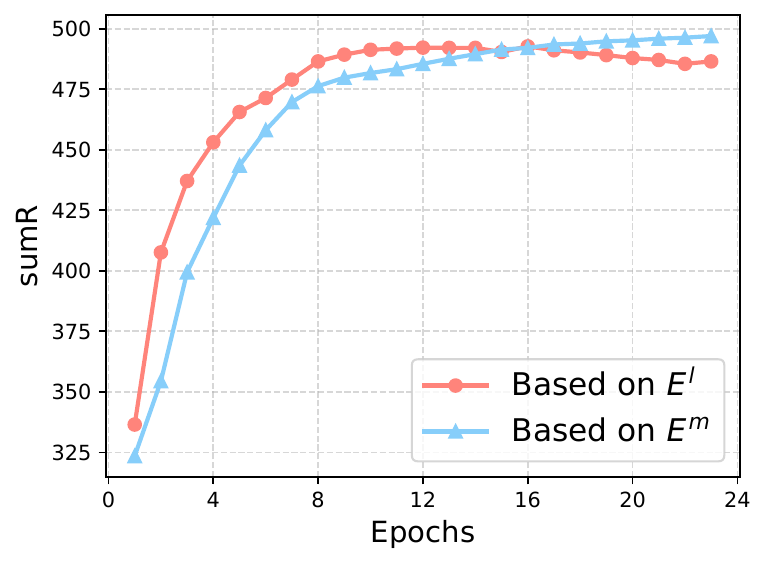}
}
\caption{(a) We train a baseline model without noise-robust learning (use \cref{eq:baseline}) and plot the losses and loss differences $\Delta$ incurred by cross-lingual alignment (red and green) and cross-modal alignment (blue and orange) during training.
(b)~Performance curves during the training on Multi30K. Confidence estimation based on $E^l$ (red) vs. confidence estimation based on $E^m$ (blue).
}\label{fig:accuracy-view} 
\end{figure}

\begin{equation}
\label{eq:evaluation_matrix}
\begin{array}{cc}
E^l = dist(f_t(S), f_t(T)) \in \mathbb{R}^{M \times M} \vspace{1ex}\\
E^m = dist(f_v(V), f_t(T)) \in \mathbb{R}^{M \times M}
\end{array}
\end{equation}

As shown in~\cref{fig:accuracy-view}~(a), the cross-lingual gap is much smaller than the cross-modal one, and the convergence rate of networks in the cross-lingual alignment is faster than that in cross-modal alignment during training.  
Therefore, confidence estimation based on $E^l$ is more accurate at the beginning of training. 
Based on this empirical finding, the transportation costs from the cross-lingual view should play a dominant role in transportation cost modeling at the preliminary stage.
Besides, considering the impact of memorization effect~\cite{arpit2017closer}, deep networks tend to first fit the clean sample pairs during an early learning stage before eventually memorizing the wrong sample pairs.
Therefore, the networks will gradually memorize noisy $(S, T)$ pairs after quickly learning cross-lingual alignment on clean data pairs. This would affect the accuracy of the confidence estimation based on $E^l$.
On the other hand, the accuracy of the confidence estimation based on $E^m$ increases as the cross-modal alignment improves progressively in the learning process.
Therefore, the networks should gradually emphasize the transportation costs from the cross-modal view in the later stage. 
To validate this idea, we conducted experiments to explore the influence of the transportation costs of different views on the performance. As~\cref{fig:accuracy-view}~(b) makes clear, the results confirm that the confidence estimation based on $E^l$ achieves better results at the early stage, while that based on $E^m$ tends to perform better at later stage. This is consistent with our assumptions.

Based on this observation, we propose a dual-view curriculum learning strategy to dynamically model the transportation costs from two views collaboratively, as illustrated in \cref{fig:example_OT}~(c). 
This strategy provides an essential dynamic curriculum where the optimization of transport costs is naturally determined by the learning state of two views. The strategy is formulated as follows:
{
\setlength\abovedisplayskip{2mm}
\setlength\belowdisplayskip{2mm}
\begin{equation}
\label{eq:curriculum_learning}
\begin{array}{cc}
E = \sigma(t) E^m + (1- \sigma (t)) E^l \\
\end{array}
\end{equation}
}
where $\sigma (t) \in [0,1]$ represents the importance of $E^m$ at time step t. At the beginning of training,
the transportation costs would focus on $E^l$, and then gradually decrease its weight until the transportation costs are dominant by $E^m$:
{
\setlength\abovedisplayskip{2mm}
\setlength\belowdisplayskip{2mm}
\begin{equation}
\label{eq:curriculum_learning_strategy}
\begin{array}{cc}
\sigma (t) = \mathbbm{1}( t \leq \tau)h( t \cdot \tau) + \mathbbm{1}( t > \tau )
   
\end{array}
\end{equation}
}
where $\tau < 1$ is an empirically-set hyper-parameter that controls the extent of $E^l$. The function $h(\cdot)$ is a non-linear curriculum to adjust the importance of each view, which can be formulated as:
{
\setlength\abovedisplayskip{1.5mm}
\setlength\belowdisplayskip{1.5mm}
\begin{equation}
    \begin{array}{cc}
        h(z) = \gamma \cdot \frac{z}{2-z}
    \end{array}
\end{equation}
}
where $\gamma$ is a hyper-parameter to control the magnitude of change. The curriculum schedule of $\sigma(t)$ ensures that the importance of $E^m$ gradually increases, and equals to 1 when $t>\tau$, which means only $E^m$ is used when $t>\tau$.

Overall, the complete procedure of confidence estimation is presented in \cref{alg:1}.

\begin{algorithm}[tb]
	\renewcommand{\algorithmicrequire}{\textbf{Input:}}
	\renewcommand{\algorithmicensure}{\textbf{Output:}}
	\caption{Confidence Estimation Algorithm}
	\label{alg:1}
	\begin{algorithmic}[1]
		\REQUIRE Noisy training dataset $\mathcal{D} = \{(V_i, S_i, T_i) \}^N_{i=1}$, max epochs $T_{max}$, iteration $I_{max}$, maximum iteration number of Sinkhorn algorithm $max\_iter$
		\ENSURE $w$ is the optimal estimated confidence of each $(V_i, S_i, T_i)$ pair.
		\FOR{$t=0,1,...,T_{max}$}  
    		\STATE Shuffle training set
    		\FOR{$n=1, ... , I_{max}$}
    		    \STATE Fetch mini-batch $\overline{\mathcal{D}}$ from $\mathcal{D}$
    		    \STATE Calculate transportation cost matrices $E_m$ and $E_l$ from two views using Eq.(8)
			    \STATE Calculate dual-view transportation costs $E$ using Eq.(9) with dual-view curriculum learning
			    \STATE Initialize $\mu^{(0)}, v^{(0)}$ as one.
		    	\WHILE{$k < max\_iter$}
			    \STATE Calculate $\mu^{(k+1)}, v^{(k+1)}$ using Eq.(6)
			    \ENDWHILE
			    \STATE Calculate optimal estimation confidence $P^{*}$ using Eq.(12)
    		\ENDFOR
    		\STATE $w = diag(P^*)$
		\ENDFOR
	    \RETURN $w$

	\end{algorithmic}  
\end{algorithm}

\subsection{Noise-aware Alignment Objective}
\label{sec:noise-aware-objective}

To promote the alignment of cross-lingual and cross-modal, we introduce pairwise alignment loss for given $\{(V_i, S_i, T_i)\}^M_{i=1}$ pairs in a mini-batch:
{
\setlength\abovedisplayskip{2.5mm}
\setlength\belowdisplayskip{2.5mm}
\begin{equation}
\label{eq:baseline}
    \begin{array}{cc}
    \mathcal{L}_{vs} = \sum^M_{i=1}  \mathcal{L}(f_v(V_i), f_t(S_i))  \vspace{1ex} \\
    \mathcal{L}_{vt} = \sum^M_{i=1}  \mathcal{L}(f_v(V_i), f_t(T_i))  \vspace{1ex} \\
    \mathcal{L}_{st} = \sum^M_{i=1}  \mathcal{L}(f_v(S_i), f_t(T_i)) 
    \end{array}
\end{equation}
}
where alignment loss function $\mathcal{L}(,)$ can be implemented by any contrastive loss. Here, we use the triplet ranking loss, which is the major loss objective for cross-modal matching tasks.
It enforces the similarity score of the matched visual-text pairs to be larger than the similarity score of the unmatched ones by a margin, formulated as:
{
\setlength\abovedisplayskip{2.5mm}
\setlength\belowdisplayskip{2.5mm}
\begin{equation}
    \begin{array}{cc}
    \mathcal{L}(l_1, l_2)  =  max(0, r + s(l_1, l_2^-) - s(l_1, l_2)) \vspace{1ex} \\ 
        +  max(0, r + s(l_2, l_1^-) - s(l_2,l_1))
    \end{array}
\end{equation}
}
where $l_1$ and $l_2$ denote the input feature vectors, $r$ indicates a margin constant and $s(\cdot)$ denotes the similarity function, \eg cosine similarity, and $l_2^-$ (or $l_1^-$) denotes a hardest negative pair for $l_1$ (or $l_2$) in the mini-batch.

Given the existence of noisy correspondences in the $(V,T)$ pairs, directly aligning them would result in the model memorizing the noisy correspondence, which would severely degrade its generalizability.
Hence, we introduce a noise-aware alignment objective, which adaptively adjusts the contribution of sample pairs based on the estimated confidence score $w$.
This objective function penalizes the noisy sample pairs less, allowing the model to focus on the more reliable samples.
{
\setlength\abovedisplayskip{2.5mm}
\setlength\belowdisplayskip{2.5mm}
\begin{equation}
    \begin{array}{cc}
    \hat{\mathcal{L}}_{vt} = \sum^M_{i=1} w_i \mathcal{L}(f_v(V_i), f_t(T_i)) 
    \end{array}
\end{equation}
}

In addition, since the main focus of our task is cross-modal retrieval, we aim to address the noisy correspondence problem in $(V, T)$ pairs.
To achieve this, we introduce cross-lingual alignment to assist the target-language encoder in learning the correct semantics from the corresponding source-language captions $S$. This helps to consistently improve the accuracy of confidence estimation based on $E^l$. However, we also need to prevent the network from overfitting to noisy $(S, T)$ pairs later in the training process. Therefore, we design a function $G(\cdot)$ that dynamically adjusts the weight of the cross-lingual objective function $L_{st}$, with the value of $G(\cdot)$ gradually decreasing in the later training stages.
{
\setlength\abovedisplayskip{1.5mm}
\setlength\belowdisplayskip{1.5mm}
\begin{equation}
    \begin{array}{cc}
        \hat{\mathcal{L}}_{st} = G(t) \mathcal{L}_{st} \vspace{1.5ex}\\
        G(t) = \frac{1}{1+ke^{(\epsilon \cdot t - \frac{1}{\tau})}}
    \end{array}
\end{equation}
}
where $k$ and $\epsilon$ are hyper-parameters. 
Finally, our objectiveness can be formulated as the combination of the above three alignment losses:
{
\setlength\abovedisplayskip{2mm}
\setlength\belowdisplayskip{2mm}
\begin{equation}
    \mathcal{L} = \hat{\mathcal{L}}_{vt} +  \hat{\mathcal{L}}_{st} + \lambda \mathcal{L}_{vs}
\end{equation}
}
where $\lambda$ determines the weight on the alignment task of images/videos and source-language captions.


\section{Experiments}

\subsection{Experimental Settings}

\textbf{Datasets.} We conduct experiments on two public multi-lingual image-text retrieval datasets:
Multi30K \cite{elliott2016multi30k} and Multi-MSCOCO, which are the multi-lingual version of Flickr30K~\cite{young2014image} and MSCOCO\cite{chen2015microsoft}, respectively, and a public multi-lingual video-text retrieval dataset VATEX~\cite{wang2019vatex}.
Noting that all non-English captions used in our training are produced by MT instead of human annotations, but human annotations are used as queries during inference.

\textit{Multi30K} is built by extending Flickr30K \cite{young2014image} from English to German, French and Czech. It contains 31,783 images and provides five captions per image in English and German and one caption per image in French and Czech. 
The human-labeled test data is provided.

\textit{Multi-MSCOCO} is extended by MSCOCO~\cite{chen2015microsoft}, and we name it Multi-MSCOCO for ease of reference. It contains 123,287 images, and each image has 5 captions. We translate the training set from English into Japanese and Chinese by resorting to MT, and follow the data split as in \cite{zhou2021uc2}.

\textit{VATEX} is a large-scale multi-lingual video dataset. Each video has 10 English captions and 10 Chinese captions to describe the video content. Note that we only use human-labeled English captions for training. Following \cite{chen2020fine}, we split the data into 25,991/1,500/1,500 as train/dev/test.

\textbf{Evaluation metrics.} Following \cite{wang2022cross}, for cross-lingual image-text retrieval, we compute the sum of all R@K~($K=1,5,10$) for both image-to-text and text-to-image retrieval, and use it (sumR) for performance comparison.
For cross-lingual video-text retrieval,
we measure rank-based performance by R@K~($K=1,5,10$) and sumR for both video-to-text and text-to-video retrieval.

\textbf{Implementation Details.}
For image encoder, we use the CLIP (ViT-B/32) \cite{radford2021learning}, a pre-trained language-image model, to extract image representations.
For video encoder, we adopt 1,024-dimensional I3D \cite{carreira2017quo} video features and use multi-layer perceptron followed by mean-pooling. 
For text encoder, we use the pre-trained mBERT-base \cite{devlin2018bert}, and take the outputs of the [CLS] token from 9-th layer as the sentence representations.

For model training, we utilize an Adam optimizer and a mini-batch size of 128. The initial learning rate is set to $2.5e-5$. We take an adjustment schedule similar to  \cite{luo2022clip4clip}.
For some hyper-parameters during training, we set the $\epsilon$, $k$ and $\lambda$ as $10$, $1$ and $0.5$ respectively. For multi30K, we set the scaling parameters of dual-view curriculum learning $\tau$ and $\gamma$ as $0.1$ and $0.2$ respectively. For Multi-MSCOCO, we set them as $0.065$ and $0.6$ respectively. For VATEX, we set them as $0.065$ and $0.1$ respectively. We use the same similarity calculation method with NRCCL during inference

\subsection{Ablation Studies}

We perform ablation studies on Multi30K to demonstrate the effectiveness of our proposed method.

\begin{table} [tb!]
\renewcommand{\arraystretch}{1.2}
\caption{Ablation study of confidence estimation based on different views on Multi30K. The methods with the checkmark (\ding{51}) introduce noise-robust learning. The metric is the sum of Recalls (sumR).  ``en", ``de", ``fr" and ``cs" indicate English, German, French and Czech, respectively.
}
\label{tab:ablation-two-view}
\centering 
\scalebox{1.0}{
\begin{tabular}{@{}l*{12}{r}c @{}}
\toprule
Cross-lingual &Cross-modal   & {\textbf{en2de}} & {\textbf{en2fr}}  & {\textbf{en2cs}}\\
\hline
\makecell{\ding{55}} &\makecell{\ding{55}} &476.6 &480.7 &470.1\\
\makecell{\ding{51}} &\makecell{\ding{55}}  &481.2 &483.3 &475.5 \\
\makecell{\ding{55}} &\makecell{\ding{51}}  &{482.6} &{484.0} &{475.6}\\
\makecell{\ding{51}}  &\makecell{\ding{51}}  &\textbf{487.4} &\textbf{490.2} &\textbf{478.1} \\
\bottomrule
\end{tabular}
}
\end{table}

\textbf{Effectiveness of dual-view collaboration.}
As shown in~\cref{tab:ablation-two-view},
the first row reports the performance of the baseline method, which is trained only using the loss of \cref{eq:baseline}. 
It assumes that all sample pairs are correctly correlated without any noise-robust designs.
Compared with the baseline, other methods with noise-robust learning have achieved performance improvement, which suggests that directly promoting the alignment of $(V,T)$ pairs will cause the neural networks to overfit the wrong supervision and degrade the performance. 
In addition, the performance is significantly improved when we combine the two views, proving the effectiveness and complementarity of our proposed dual-view collaboration.

\begin{figure}[tb!]
\centering\includegraphics[width=0.8\columnwidth]{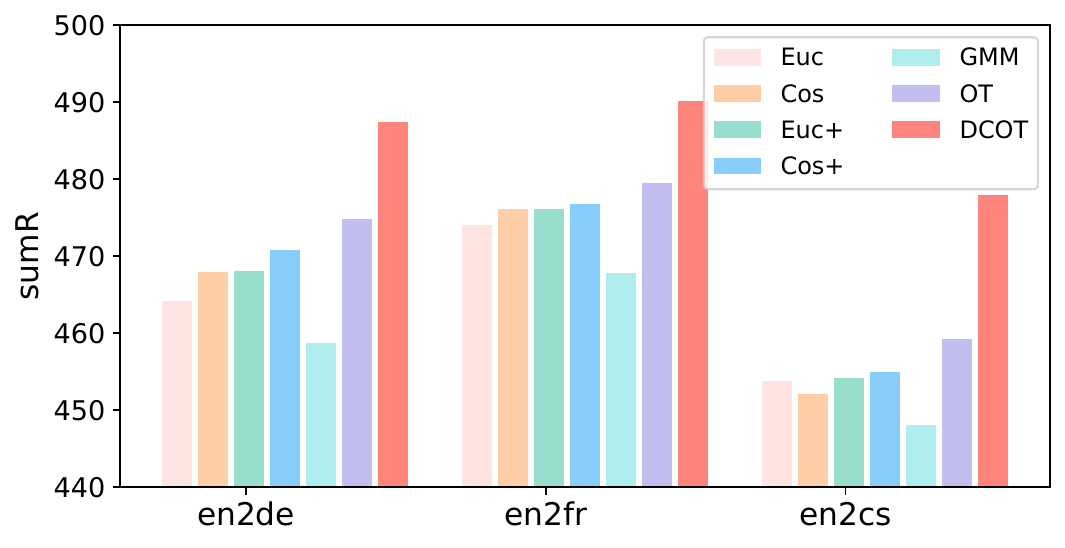}
\caption{
Performance comparison of P2P-based methods, GMM-based method and our proposed OT-based confidence estimation method on Multi30K. 
``Euc" and ``Cos" denote the Euclidean and Cosine distance, respectively. The symbol ``+"  denotes that a softmax operation is applied after the distance calculation.
}\label{fig:ablation-ot}
\end{figure}

\textbf{Effectiveness of OT-based confidence estimation.}
To validate the effectiveness of the confidence estimation using OT, 
we compare it with the counterparts using P2P calculation without context (\ie Euc, Cos, Euc+, and Cos+) and GMM in confidence estimation.
As shown in~\cref{fig:ablation-ot}, the OT-based methods achieve significant advantages in all languages.
The P2P-based methods suffer from the lack of contextual information, which cannot provide accurate confidence estimation. In contrast, our OT-based algorithm could estimate confidence based on the principle of minimum global costs, by taking the mutual relation between the samples into account. 
Moreover, the performance of the GMM-based method is severely degraded when the noise distribution deviates from the Gaussian distribution, given that the method has strong constraints on the data distribution. Compared with it, our OT-based method does not require any assumptions about the data distribution, making it more versatile.
Additionally, DCOT beats the other methods by a large margin, demonstrating the importance of collaborative effort between the two views, especially in the low-resource languages (\eg Czech in en2cs).
The promising results validate that our proposed confidence estimation algorithm can achieve more accurate confidence and greatly enhance the robustness of the model to noise.

\begin{figure}[tb!]
\centering
\subfigure[Basic Model]{
\label{fig:curve-a}
\includegraphics[width=0.43\columnwidth]{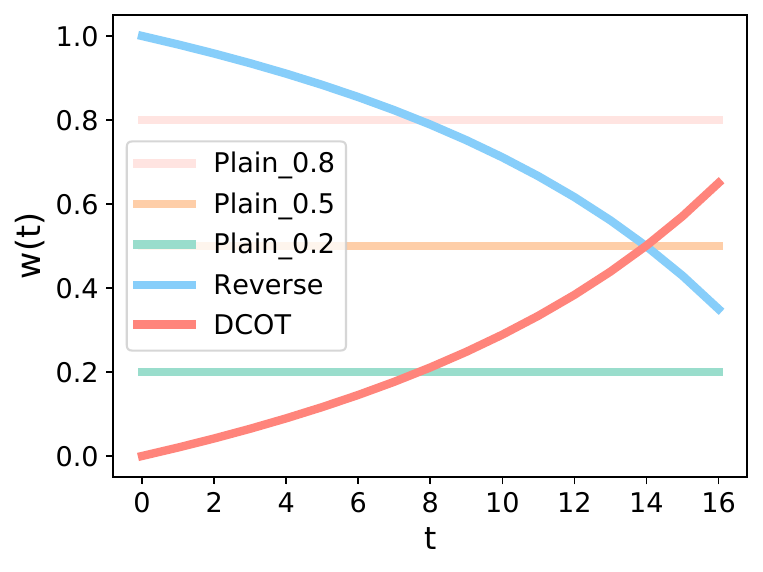}
}
\quad
\subfigure[Ours]{
\label{fig:performance-b}
\includegraphics[width=0.43\columnwidth]{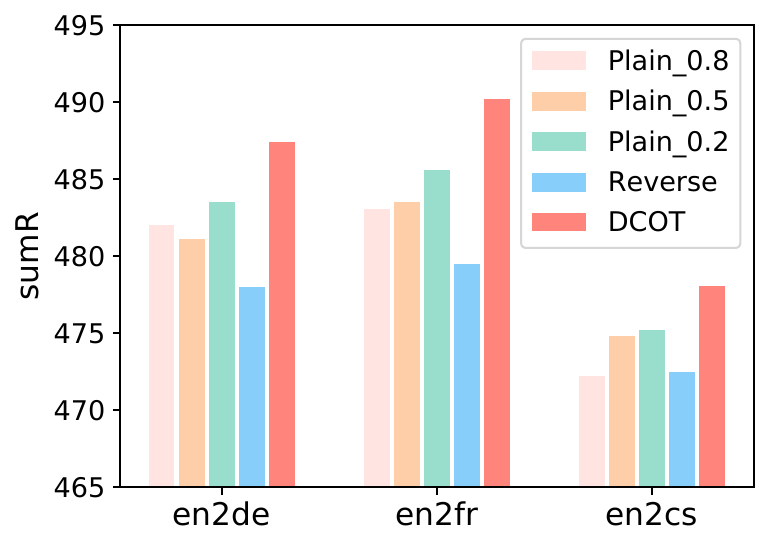}
}
\caption{(a) Examples of various curriculum schedules. 
We only show the curve in the cross-modal view for ease of representation. The curve in the cross-lingual view corresponds to $1-w(t)$.
(b) Performance comparison of various curriculum schedules on Multi30K.
}\label{fig:ablation-schedules}
\end{figure}

\textbf{Influence of curriculum schedule.}
To perform an in-depth study on dual-view curriculum learning, we conduct experiments with various curriculum schedules shown in~\cref{fig:ablation-schedules}~(a) and the results are displayed in~\cref{fig:ablation-schedules}~(b). As we can see, the reverse schedule performs the worst.
The reason lies in that the collaboration process of two views exactly deviates from the relative accuracy of the confidence estimation based on the two views, leading to a significant performance drop.
For plain schedules, the importance of each view remains fixed throughout the training. 
In contrast, our proposed dynamic schedule adjusts the importance of each view according to their learning state, which has a significant superiority. 
Thus, designing the appropriate schedule is crucial in improving the accuracy of the confidence estimation.

\subsection{Comparison with State-of-the-Arts}

\subsubsection{Cross-lingual Image-Text Retrieval}
For cross-lingual image-text retrieval, 
we compare four state-of-the-art (SOTA) methods, including M$^3$P \cite{ni2021m3p}, UC$^{2}$ \cite{zhou2021uc2}, CCLM \cite{zeng2022cross}, and NRCCR \cite{wang2022cross}.
Among them, M$^3$P, UC$^{2}$, and CCLM are all pre-trained on the large-scale vision-language corpus, while NRCCR is the robust learning method against noisy correspondence.
For a fair comparison, we compare DCOT to CCLM with the dual-stream structure, as the dual-stream models are more suitable for large-scale retrieval.

\begin{table} [tb!]
\renewcommand{\arraystretch}{1.2}
\caption{Performance comparison of cross-lingual image-text retrieval on Multi30K (the source language is English and the target language is non-English). 
The scores are the sum of all Recalls (sumR).
The symbol asterisk (*) indicates that model was pre-trained on large-scale datasets, e.g., CC3M and its MT version.
}
\label{tab:sota-multi30k}
\centering 
\scalebox{1.0}{
\begin{tabular}{@{}l*{12}{r}c @{}}
\toprule
\textbf{Method} & \makecell{\textbf{Backbone (\#parameters)}} & {\textbf{en2de}} & {\textbf{en2fr}}  & {\textbf{en2cs}}\\
\hline
M$^3$P \cite{ni2021m3p}* &\makecell{XLMR-large (560M)} &351.0 &276.0 &220.8\\
UC$^2$ \cite{zhou2021uc2}*  &\makecell{XLMR-base (278M)} &449.4 &444.0 &407.4\\
CCLM  \cite{zeng2022cross}* 
&\makecell{XLMR-large (560M)}  &\underline{503.4} &490.6 &481.6 \\
NRCCR \cite{wang2022cross} &\makecell{mBERT (170M)} &{480.6} &{482.1} &{467.1}\\
DCOT(ours)  &\makecell{mBERT (170M)} &{494.9} &\underline{495.3} &\underline{481.8} \\
CCLM+ours*  &\makecell{XLMR-large (560M)} &\textbf{515.2} &\textbf{518.7} &\textbf{512.1} \\
\bottomrule
\end{tabular}
 }
\end{table}

\textbf{Comparisons on Multi30K.}
\cref{tab:sota-multi30k} summarizes the performance comparison on Multi30K.
Without pre-training and using a more lightweight backbone, 
DCOT outperforms the large-scale pre-trained model M$^3$P and UC$^2$ by a large margin, and achieves comparable performance to CCLM. 
Moreover, the sumR scores of DCOT is $2.2\%$, $2.7\%$ and $3.1\%$ higher than noise-robust learning baseline NRCCR on three languages, respectively.
As NRCCR obtains pseudo supervision signals by calculating point-to-point correspondence of $(V,T)$ pairs, it does not take the mutual relation between features into account. By contrast, our DCOT models the confidence estimation from a contextual perspective. The results demonstrate that our proposed contextual modeling is more beneficial for noisy correspondence learning.

Recall that our proposed dual-view curricular optimal transport is orthogonal to cross-lingual image-text similarity learning.
In this experiment, we evaluate its expansibility to cross-lingual image-text baselines. 
Specifically, we employ our dual-view curricular optimal transport on the recent state-of-the-art cross-lingual image-text method CCLM~\cite{zeng2022cross} during finetuning.
As shown in \cref{tab:sota-multi30k}, applying our proposed dual-view curricular optimal transport for noise correspondence learning consistently brings improvement in all languages.
Note that CCLM does not consider the noise of training data.
These results not only demonstrate the  good expansibility of our method to cross-lingual image-text similarity learning, but also further verify the effectiveness of our noise correspondence learning.

\begin{table} [tb!]
\renewcommand{\arraystretch}{1.2}
\caption{Performance comparison of cross-lingual image-text retrieval on Multi-MSCOCO. ``zh" and ``ja" indicate the Chinese and Japanese, respectively.
}
\label{tab:sota-muti-mscoco}
\centering 
\scalebox{1.0}{
\begin{tabular}{@{}l*{12}{r}c @{}}
\toprule
\textbf{Method}  & \makecell{\textbf{Backbone (\#parameters)}} & {\textbf{en2zh}} & {\textbf{en2ja}}\\
\hline
M$^3$P \cite{ni2021m3p}* &\makecell{XLMR-large (560M)} &322.8 &336.0\\
UC$^2$ \cite{zhou2021uc2}* &\makecell{XLMR-base (278M)} &492.0 &430.2 \\
CCLM  \cite{zeng2022cross}*  
&\makecell{XLMR-large (560M)} &511.2 &{496.4} \\
\hline

NRCCR \cite{wang2022cross}  &\makecell{mBERT (170M)} &{512.4} &{507.0} \\
DCOT(ours) &\makecell{mBERT (170M)} &\underline{521.5} &\underline{515.3} \\
CCLM+ours* &\makecell{XLMR-large (560M)} &\textbf{535.6} &\textbf{536.2} \\
\bottomrule
\end{tabular}
}
\end{table}

\textbf{Comparisons on Multi-MSCOCO.}
\cref{tab:sota-muti-mscoco} reports the experimental results on Multi-MSCOCO. 
DCOT significantly outperforms large-scale pre-trained models that do not consider the noisy correspondence problem.
Compared to NRCCR, DCOT still has a huge advantage, with a $1.7\%$ and $1.6\%$ improvement in terms of sumR.
Notably, 
compared to German and French in Multi30K, Chinese and Japanese in Multi-MSCOCO exhibit significant structural differences from English, making them more susceptible to noise during the translation process.
Thus, Multi-MSCOCO is more challenging than Multi30K.
The better performance of DCOT on Multi-MSCOCO than on Multi30K further demonstrates its superior robustness against noise.
Besides, applying our method to CCLM also achieves a significant performance gain, showing that our method is compatible with popular pre-training models.

\begin{table} [tb!]
\renewcommand{\arraystretch}{1.2}
\caption{Performance comparison of cross-lingual video-text retrieval on VATEX (the source language is English and the target language is Chinese).
Symbol asterisk (*) indicates the model is pre-trained on Multi-HowTo100M \cite{huang2021multilingual}.}
\label{tab:sota-vatex}
\centering 
\scalebox{0.85}{
\begin{tabular}{@{}l*{12}{r}c @{}}
\toprule
\multirow{2}{*}{\textbf{Method}}   & \multicolumn{3}{c}{\textbf{Text $\rightarrow$ Video}} &  & \multicolumn{3}{c}{\textbf{Video $\rightarrow$ Text}}  & \multirow{2}{*}{\textbf{sumR}}\\ 
\cmidrule{2-4} \cmidrule{6-8}
& R@1  & R@5  & R@10  & & R@1 & R@5 & R@1   \\ \hline
MMP \cite{huang2021multilingual}  & 23.9  & 55.1  & 67.8  &  & - & {-} & {-} & {-} \\
MMP \cite{huang2021multilingual} *  & 29.7  & 63.2  & 75.5  &  & {-} & {-} & {-} & {-}\\ 
NRCCR \cite{wang2022cross}  & 30.4 & 65.0 & 75.1 &   & 40.6 & 72.7 & 80.9  & 364.7\\ 
{DCOT}  & \textbf{31.4} & \textbf{66.3} & \textbf{76.8} &  & \textbf{46.0} & \textbf{76.3}  & \textbf{84.8} & \textbf{381.8}\\
\bottomrule
\end{tabular}
 }
\end{table}

\subsubsection{Cross-lingual Video-Text Retrieval}

For cross-lingual video-text retrieval, we compare our model with two SOTA methods, MMP \cite{huang2021multilingual} and NRCCR \cite{wang2022cross}. Among them, MMP is pre-trained on Multi-HowTo100M (the multi-lingual version of HowTo100M \cite{miech19howto100m}), and NRCCR is the robust learning method against noise introduced by MT.
As shown in~\cref{tab:sota-vatex}, DCOT demonstrates superior performance compared to large-scale pre-trained model MMP, which verifies the benefit of mitigating the noisy correspondence problem. Compared to the best baseline NRCCR, DCOT outperforms it by $4.7\%$ in terms of sumR. From the results, one could see that our DCOT achieves excellent results, with the best results for cross-lingual video-text retrieval.

\begin{figure*}[tb!]
\centering
\subfigure[en2de]{
\label{fig:a}
\includegraphics[width=0.52\columnwidth]{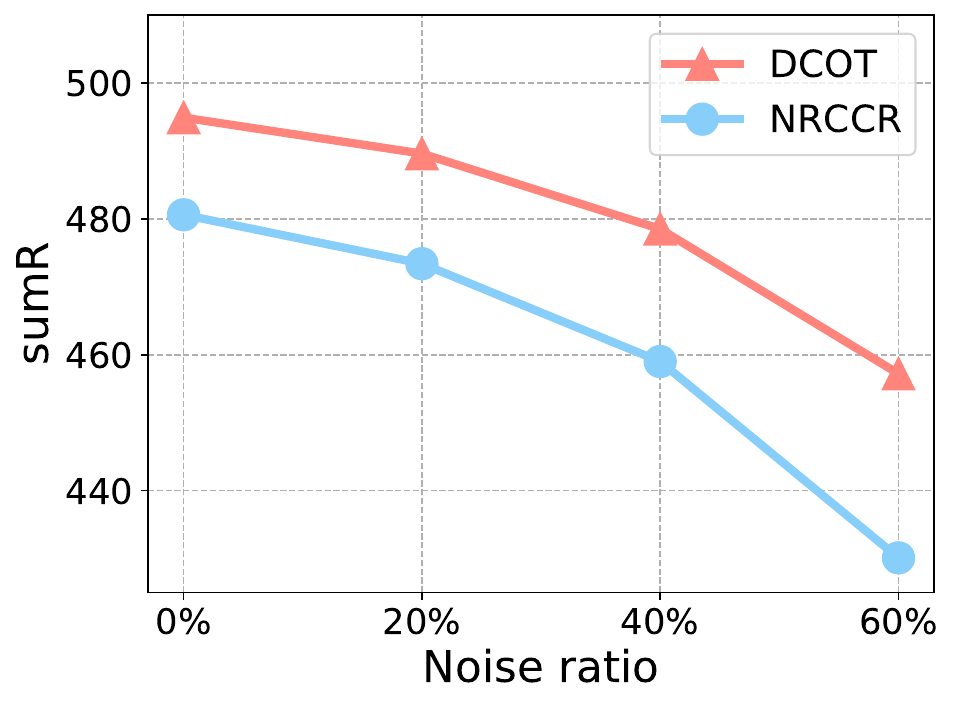}
}
\subfigure[en2fr]{
\label{fig:b}
\includegraphics[width=0.52\columnwidth]{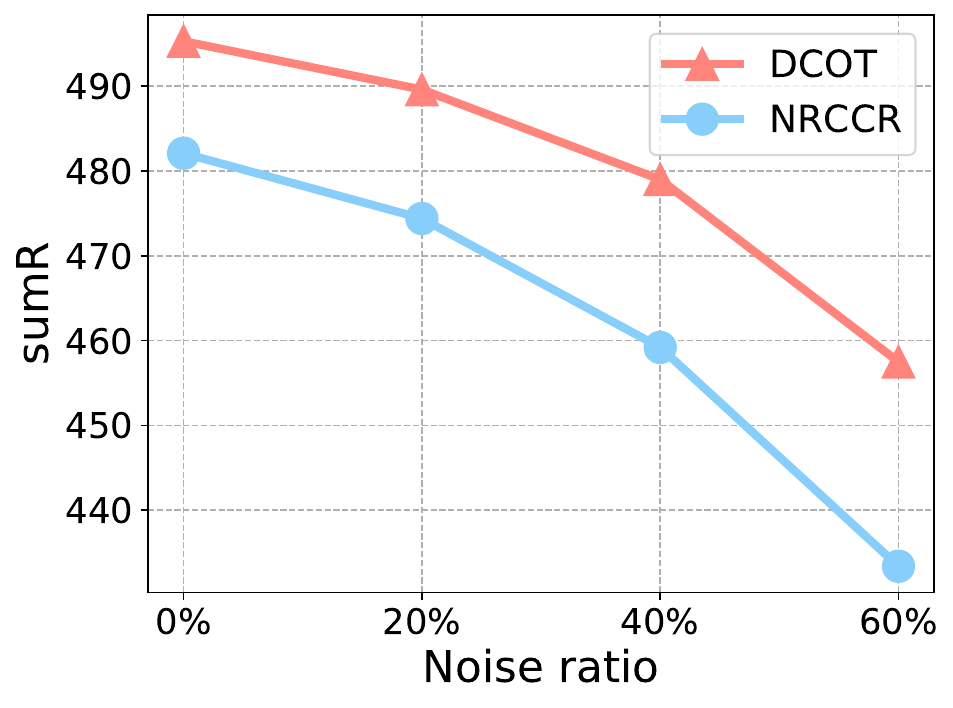}
}
\subfigure[en2cs]{
\label{fig:c}
\includegraphics[width=0.52\columnwidth]{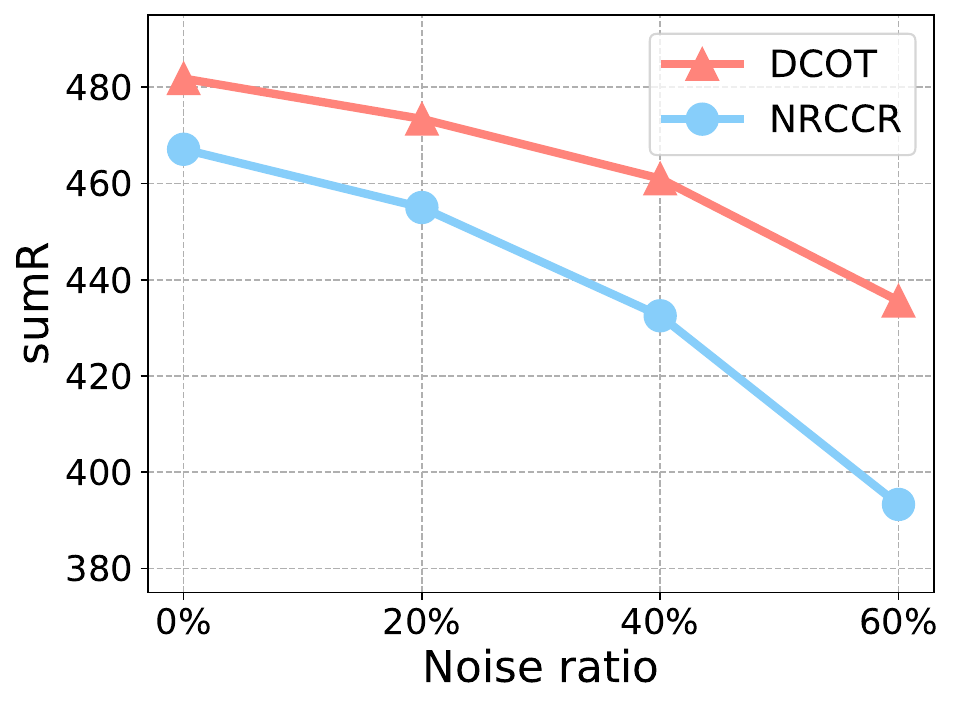}
}
\caption{Performance comparison with different noise ratios on Multi30K. The noise is injected by switching the correspondence of $(V, T)$ pairs artificially, where ``0\%" means no noise is added artificially. 
}\label{fig:noise} 
\end{figure*}

\begin{table} [tb!]
\renewcommand{\arraystretch}{1.2}
\caption{Zero-shot results on Multi30K. ``-MT" indicates the multi-lingual version. Note that CC3M and CC3M-MT both contain 3,346,732 image-text pairs, while Multi-MSCOCO only contains 616,435 image-text pairs. The results suggest that noise-robust learning can alleviate the dependence on large-scale training data.}
\label{tab:sota-zero-shot}
\centering 
\scalebox{1.0}{
\begin{tabular}{@{}l*{12}{r}c @{}}
\toprule
\textbf{Method} &\makecell{\textbf{Training Data}} & {\textbf{en2de}} & {\textbf{en2fr}}  & {\textbf{en2cs}}\\
\hline
M$^3$P \cite{ni2021m3p} &\makecell{CC3M + Wikipedia}  &220.8 &162.6 &122.4\\
UC$^2$ \cite{zhou2021uc2} &\makecell{CC3M-MT} &375.0 &362.4 &330.6 \\
CCLM  \cite{zeng2022cross}  &\makecell{CC3M-MT} &409.5 & 384.4 &375.3 \\
NRCCR \cite{wang2022cross} &\makecell{Multi-MSCOCO} &{448.7} &{433.8} &{411.2}\\
DCOT &\makecell{Multi-MSCOCO} &\textbf{458.9} &\textbf{445.3} &\textbf{424.2} \\
\bottomrule
\end{tabular}
}
\end{table}

\subsection{Generalization Analysis on Out-of-domain Data}

In this section, we evaluate the generalization capability on out-of-domain data of our proposed method.
In \cref{tab:sota-zero-shot}, we report results on cross-lingual image-text retrieval under the zero-shot setting, where training data and testing data are from different domains. 
Compared with large-scale pre-trained model M$^3$P and CCLM, DCOT and NRCCR obtain superior results with fewer training data.
Recall that both DCOT and NRCCR are trained in noise-robust manners.
The results suggest that noise-robust learning can alleviate the dependence on large-scale training data and verifies the essential of noise-robust learning.
In addition, with the same training data, our method consistently outperforms the noise-robust learning method NRCCR by a significant margin.
This result verifies our proposed method DCOT has better cross-lingual transfer ability under the noisy scenario.

\begin{figure}[tb!]
\centering
\subfigure[origin]{
\label{fig:origin}
\includegraphics[width=0.45\columnwidth]{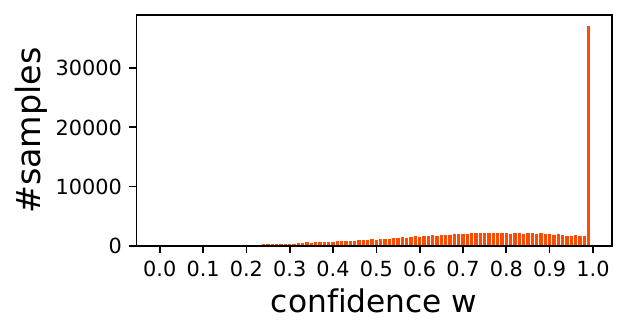}
}
\subfigure[$20\%$ noise]{
\label{fig:b}
\includegraphics[width=0.45\columnwidth]{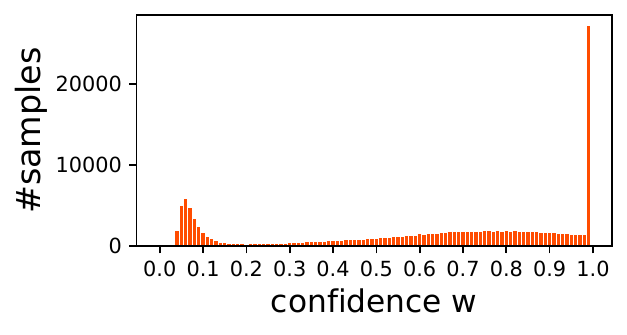}
}
\\
\subfigure[$40\%$ noise]{
\label{fig:c}
\includegraphics[width=0.45\columnwidth]{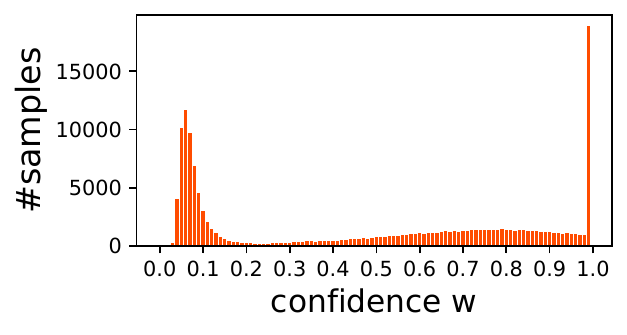}
}
\subfigure[$60\%$ noise]{
\label{fig:c}
\includegraphics[width=0.45\columnwidth]{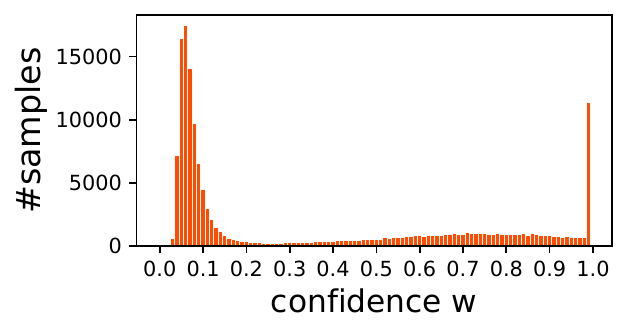}
}
\caption{Distribution of estimated confidence $w$ under different noise ratios on French of Multi30K. 
}\label{fig:confidence} 
\end{figure}

\subsection{Robustness Analysis}

To investigate the robustness of our proposed model, we conduct experiments with four different noise ratios on Multi30K. 
We compare our method with NRCCR which is the only noise-robust learning method for CCR.
Specifically, we corrupt the training data by switching the correspondence of $(V, T)$ pairs of some random instances based on a noise rate parameter. 
The higher the noise rates, the more serious the noisy correspondence problems become. 
The performance curves on two languages with the artificial noise ratios $[0, 0.2, 0.4, 0.6]$ are shown in~\cref{fig:noise}.
Our DCOT consistently performs better than NRCCR, and the performance gap between DCOT and NRCCR becomes larger as the noise rate increases. 
The results clearly show that DCOT performs more stability than NRCCR.

\subsection{Visualization Analysis}
\subsubsection{Confidence Visualization}
To further investigate the effectiveness of confidence estimation in our method, we carry out experiments by visualizing the per-sample confidence distribution under the different noise ratios. 
Specifically, we corrupt the training data by randomly switching the correspondence of $(V, T)$ pairs of some instances using a specific noise rate.
As the noise rates increased, the noisy correspondence problems became more severe.
From \cref{fig:confidence}, we can observe that the number of low-confidence samples increases with the noise ratio, which verifies the estimated confidence can reflect the magnitude of the noise.

\subsubsection{Retrieval Visualization}

\begin{figure*}[tb!]
\centering\includegraphics[width=2.0\columnwidth]{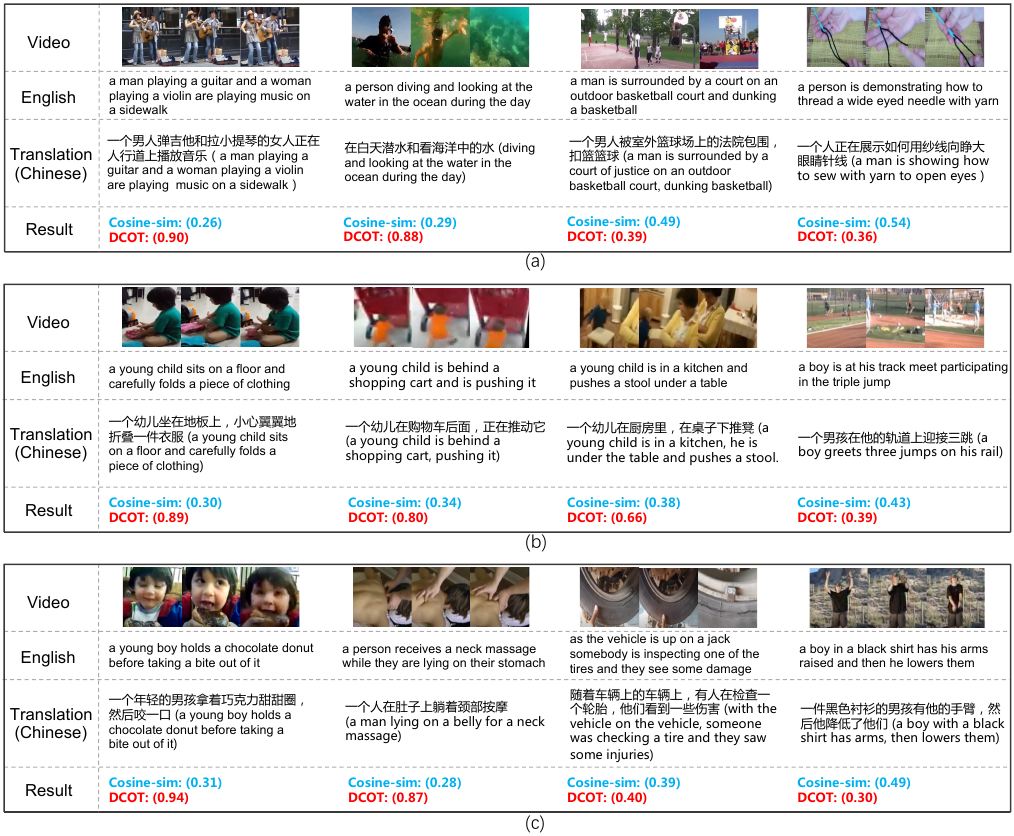}
\vspace{-2mm}
\caption{
Qualitative results of noisy correspondence learning on VATEX. 
We compare our proposed DCOT to the cosine similarity that is a point-to-point correspondence calculation method without considering the context.
Our DCOT with context modeling achieves more reasonable confidence estimation results.
}\label{fig:qualitative-result}
\vspace{-3mm}
\end{figure*}

In \cref{fig:qualitative-result}, we present the qualitative results on VATEX.
From the results, we could observe that our DCOT could provide a more reasonable confidence estimation between sample pairs than the point-to-point counterpart that directly utilizes the cosine similarity to measure the correspondence without considering the context. 
Take \cref{fig:qualitative-result}~(a) as an example.
The translation of the first example is significantly more accurate than the last one, and our DCOT accordingly gives the highest score to the first one. In contrast, the cosine similarity based counterpart outputs the lowest score for this example.
Moreover, we also observe that the cosine similarity scores vary in a very small range, which hardly reflects the quality of video-target language caption pairs. For the second example in  \cref{fig:qualitative-result} (c), our method fails to estimate confidence accurately. We assume that this is because the confidence estimation of our approach is based on global features, which capture the semantic information of a neck massage. However, it still has limitations in fine-grained information confidence estimation (\eg lying on a belly).

\section{Conclusion}

In this paper, we focus on the noisy correspondence problem in CCR. We propose a novel method, called dual-view curricular optimal transport (DCOT), which formulates the noisy correspondence learning in CCR as an optimal transport problem. 
We estimate the confidence of the correlated sample pair from both the cross-lingual and cross-modal views and design a dual-view collaborative curriculum learning strategy to model the transportation costs dynamically according to the learning state of the two views. 
Additionally, we adjust the contribution of each data pair based on the estimated confidence to avoid network overfitting to the noisy sample pairs.
Extensive experiments demonstrate the robustness and effectiveness of our method against noise introduced by MT. 
In future work, we plan to conduct an in-depth study of the impact of different levels of noise introduced by MT and explore how to adaptively adjust the curriculum schedule for different languages.

\bibliography{main}
\bibliographystyle{IEEEtran}
\newpage

\vfill

\end{document}